\documentclass{INTERSPEECH2023}
\interspeechcameraready 

\usepackage{xcolor} % textcolor
\usepackage{algorithm} % algorithm
\usepackage{algpseudocode} % algorithmic
\usepackage{multirow} % multirow/multicol in table
\usepackage{hyperref} % href
\usepackage{cleveref} % cref
\usepackage{cite}
\usepackage{xfrac}
\usepackage{diagbox} % backslashed box in table

\newcommand{\STAB}[1]{\begin{tabular}{@{}c@{}}#1\end{tabular}}

\creflabelformat{equation}{#2\textup{#1}#3} % ex) eq. 1

\title{Unsupervised Speech Representation Pooling Using Vector Quantization}
\name{Jeongkyun Park$^*$\thanks{* Equal contribution.}, Kwanghee Choi$^*$, Hyunjun Heo$^*$, Hyung-Min Park}
\address{Sogang University}
\email{\{park32323,juice500,hteconn,hpark\}@sogang.ac.kr}

\begin{document}

\maketitle
\begin{abstract}\label{sec:abstract}
With the advent of general-purpose speech representations from large-scale self-supervised models, applying a single model to multiple downstream tasks is becoming a de-facto approach. However, the pooling problem remains; the length of speech representations is inherently variable. The naive average pooling is often used, even though it ignores the characteristics of speech, such as differently lengthed phonemes. Hence, we design a novel pooling method to squash acoustically similar representations via vector quantization, which does not require additional training, unlike attention-based pooling. Further, we evaluate various unsupervised pooling methods on various self-supervised models. We gather diverse methods scattered around speech and text to evaluate on various tasks: keyword spotting, speaker identification, intent classification, and emotion recognition. Finally, we quantitatively and qualitatively analyze our method, comparing it with supervised pooling methods.
\end{abstract}
\noindent\textbf{Index Terms}: pooling, vector quantization, speech representations, self-supervised models, wav2vec 2.0
\section{Introduction}\label{sec:intro}
% P1: Advent of self-supervised large-scale model, general speech embeddings
wav2vec 2.0 \cite{baevski2020wav2vec} has swept the speech processing community like a storm with the surprising effectiveness of transformer-based self-supervised models \cite{yang2021superb}.
Self-supervision leverages large quantities of unlabeled data, which is much easier to obtain than labeled data.
Many task designs exist, but one of the most common is to mask the input and train the neural network to reconstruct the original input based on the surrounding information \cite{kenton2019bert,baevski2020wav2vec}.
The task definition enables the neural network to model the input distribution so that it can yield a general representation useful for various downstream tasks.

% P2: pooling is important
After building a general self-supervised model, supervised training ensues to fine-tune the model to each specific need.
One often attaches the penultimate fully connected layer to the frozen shared model, acting as the prediction head \cite{yang2021superb}.
However, input lengths often vary for both speech and text, a fundamental characteristic of sequential data.
Transformer's output representation length linearly increases as the input size gets longer, while the final prediction head requires the fixed-size input representation.
It raises the problem of representation pooling; given a variable number of sequential representations, one has to summarize the representation to have a fixed size.

% P3: NLP Pooling and their limitations
The pooling task is often addressed as the unsupervised or paraphrastic sentence embedding problem in the NLP literature \cite{wieting2015towards,arora2017simple}, especially after the success of word embeddings on various tasks.
It aims to yield a general-purpose sentence embedding independent of each downstream task.
By starting from simply averaging the word representations \cite{wieting2015towards}, many unparameterized methods were introduced, such as considering the word frequency while averaging \cite{arora2017simple}, whitening the representations \cite{huang2021whiteningbert,su2021whitening}, or utilizing the singular value transformation \cite{yan2022addressing}.
However, existing methods are often dependent on the underlying tokenizer, making it nontrivial to apply to speech.
Also, simply averaging the speech representations poses a fundamental problem; each phone has a different length, so vowels will be overrepresented compared to consonants.

% P4: SLP Pooling and their limitations
In the meantime, many pooling methods have been introduced for summarizing the frame-wise speech representations, notably for speaker recognition.
For example, statistics pooling (SP) \cite{snyder2016deep,snyder2017deep} concatenates the first- and second-order statistics of representations.
Also, the attention layer is often used to obtain which representation matters more for each downstream task \cite{okabe2018attentive,safari2020self,wu2020vector}.
However, there has been limited interest in developing an unsupervised way of pooling.
Modern methods, except for SP, are often heavily parameterized, requiring labeled samples.
In contrast, unsupervised pooling removes the training procedure altogether, being more closer to the idea of general speech embedding.

% P5: we shed light on vq again... for speech-friendly unsupervised pooling method.
To avoid parameterization while being speech-friendly, we shed light on vector quantization (VQ) to substitute the tokenizer.
Decades of research have been conducted on VQ \cite{soong1987report}, where it crept into the modern self-supervised models, such as vq-wav2vec \cite{jegou2010product,baevski2020vq}.
VQ focuses on translating the real-valued vectors into countable indices, segmenting the latent embedding space with well-chosen centroids.
Given that essential components comprising speech, such as frequency, amplitude, and formants, are embedded inside the model representation\cite{choi2022opening}, we expect the clusters to successfully gather similar phone representations to have the same cluster indices.
We design various intuitive approaches that utilize VQ to effectively summarize the speech representations, and experimentally demonstrate its effectiveness in a wide range of settings.

% P6: further, we devise a benchmark for comparing pooling methods, evaluation scheme, task...
To further boost the research on unsupervised pooling methods, we devise a benchmark that evaluates various pooling methods across multiple backbone networks and a wide range of tasks, namely, keyword spotting \cite{warden2018speech}, speaker identification \cite{nagrani2017voxceleb}, intent classification \cite{lugosch2019speech}, and emotion recognition \cite{busso2008iemocap}.
To avoid supervision altogether, we closely follow the recent contrastive learning literature \cite{chen2020simple,choi2022combating} to evaluate the effectiveness of the pooled representation directly via the nearest-neighbor approach \cite{bernhardsson2017annoy}.
Also, we modify the existing unsupervised sentence embedding methods to compare their downstream performance.
We further conduct various analyses that accurately depict the effectiveness and behaviors of our method by comparing with the parameterized methods via supervised learning, visualizing the weights directly, and exploring different settings for benchmark evaluation.

% P7. sectionwise summary / method, benchmark, results, further analysis
% pass
\section{Method}\label{sec:method}
In this section, we briefly overview the vector quantization method used in vq-wav2vec \cite{baevski2020vq} and wav2vec 2.0 \cite{baevski2020wav2vec,babu2022xls-r} in \Cref{ssec:vq}.
Then, we discuss various ways to utilize the quantization approach.
\Cref{ssec:vq_merge} covers the idea of merging a similar representation, while \Cref{ssec:vq_prior} utilizes the quantized results to estimate the prior probability of each representation.

\subsection{Preliminaries: vector quantization}\label{ssec:vq}
We briefly cover the internals of wav2vec 2.0 \cite{baevski2020wav2vec} and its vector quantizer module before introducing our method.
Let us denote the raw speech input as $\mathbf{x}$, latent speech representations from the 1D convolutional features as $\mathbf{Z}$, context representations from the transformer as $\mathbf{C}$, and the quantized representation as $\mathbf{Q}$:
\begin{align}
    \mathbf{Z} &= \texttt{CNN}(\mathbf{x}) \in \mathbb{R}^{T \times F}, \\
    \mathbf{C} &= \texttt{transformer}(\mathbf{Z}) \in \mathbb{R}^{T \times F}. \\
    \mathbf{Q} &= \texttt{quantizer}(\mathbf{Z}) = \{q_t^g \} \in [V]^{T \times G},
\end{align}
where $G$ is the number of codebook groups, where sharing the codebooks often yield comparable performance with a much better efficiency \cite{baevski2020vq}.
We denote an integer $i$ between $1 \leq i \leq V$ as $[V]$.
Each representation of the $t$-th frame is represented as $G$ integers, where each integer is $1 \leq q_t^g \leq V$, where $V$ is the number of centroids.
Hence, for each frame-wise speech representation $\mathbf{Z}_t \in \mathbb{R}^F$ where $t$ is the frame index, the quantizer maps $\mathbf{Z}_t$ to the $G$-dimensional integers $(q_t^1, q_t^2, \cdots, q_t^G)$.
The quantized results are designed to map the nearby representations to have the same indices, such that $\texttt{quantizer}(\mathbf{Z}) = \texttt{quantizer}(\mathbf{Z} + \epsilon)$ for a negligible acoustic difference $\epsilon$.

\subsection{Squashing via quantized results}\label{ssec:vq_merge}
For the ease of notation, we regard pooling as the weighted sum of frame representations $\mathbf{C}_t$ with a scalar $w_t$:
\begin{equation}
    \Bar{\mathbf{C}} = \textstyle\sum_{t=1}^{t=T} \zeta w_t \mathbf{C}_t.
\end{equation}
where the normalizing constant $\zeta$ is used to ensure that the sum of weights equals $1$, \textit{i.e.}, $\textstyle\sum_{t=1}^{t=T} \zeta w_t = 1$.

For instance, for average pooling (AP), $w_t=1$ and $\zeta = 1/T$.
We observe that the AP regards every frame to be equally important.
However, the length of speech signals is not proportional to their importance.
For instance, even if there is a long silence, the AP will average them, and the resulting representation will be highly biased towards silence representation.
Also, vowels and consonants have inherently different lengths; hence the representation will be fitted towards vowels more.

\begin{figure}[t]
\vspace{-1em}
\centering
\includegraphics[width=1.0\columnwidth]{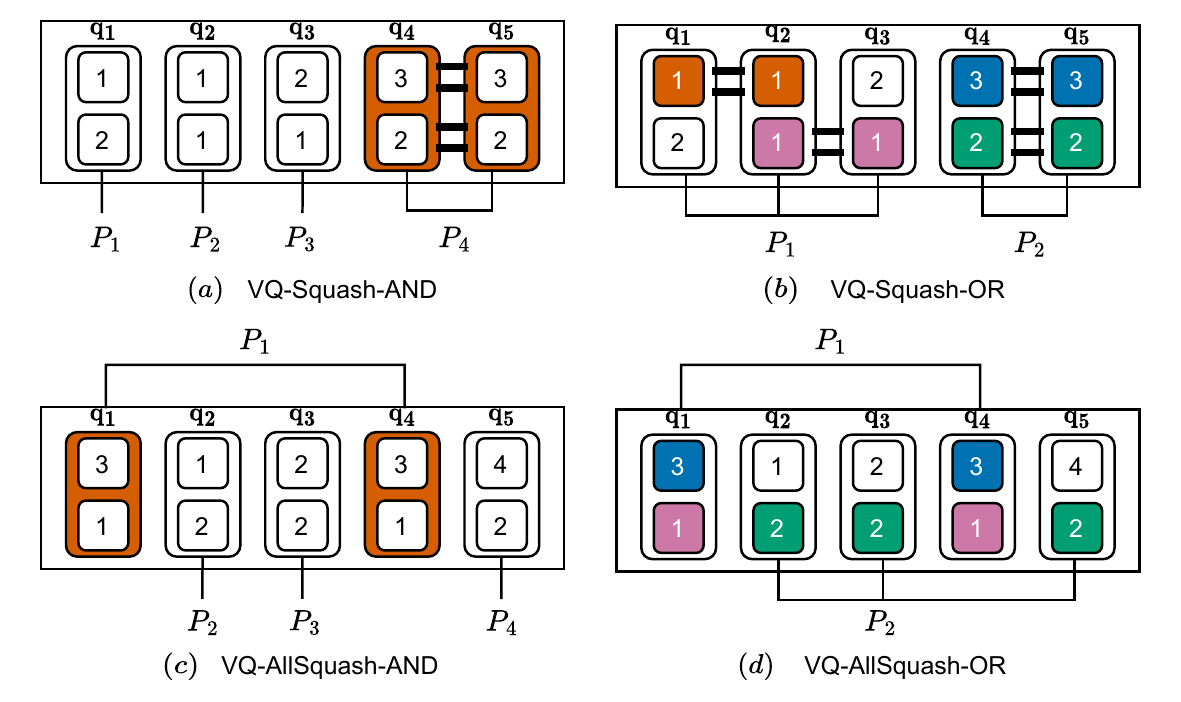}
\vspace{-2em}
\caption{
Partition Methods for Quantized Representations.
}
\vspace{-1em}
\label{fig:vq}
\end{figure}

Hence, we develop a novel pooling approach based on VQ, which clusters similar frames into each partition, borrowing the idea from greedy decoding of Connectionist Temporal Classification (CTC) acoustic models \cite{zenkel2017comparison}.
By averaging with respect to partitions, the method effectively penalizes the representations within the larger partition:
\begin{equation}
    \Bar{\mathbf{C}}_\text{Squash} = \frac{1}{|\mathcal{S}(\mathbf{Q} )|} \sum_{P \in \mathcal{S}(\mathbf{Q})} \frac{1}{|P|} \sum_{t \in P} \mathbf{C}_t.
\end{equation}
$\mathcal{S}: [V]^{T \times G}\rightarrow 2^{[V]^{T \times G}}$ is the squashing function, where we write the superset of a set $A$ as $2^A$.
The squash function returns a partition of the frame indices, where each partition is constructed by squashing similar representations.
The inner fraction exists for averaging within a single partition that contains similar representations ($\sfrac{1}{|P|}$), where the outer fraction ($\sfrac{1}{|\mathcal{S}(\mathbb{Q})|}$) exists for averaging across partitions.
We explore two perspectives on squashing, summarized in \Cref{fig:vq}.

Firstly, the equality of two representations at different steps $\mathbf{C}_i$ and $\mathbf{C}_j$ can be defined in two ways.
We can check all the indices of $\mathbf{C}_i$ and $\mathbf{C}_j$ (\texttt{AND}), or allow partial matches (\texttt{OR}).
Compared to $\texttt{AND}$, $\texttt{OR}$ aims to merge more neighboring samples by using each quantized group while tolerating some errors.

Further, one has to define which temporal frames should be merged.
We develop a strategy \texttt{Squash}, which penalizes the features that occur repeatedly in long sequences, similar to CTC, where neighboring equal values are squashed.
On the other hand, we observe that AP works in an orderless manner.
Hence one can extend the definition of \texttt{Squash} to be averaging all the occurrences, penalizing the features that occur more often than others.
We call this strategy \texttt{AllSquash}.

\subsection{Prior probability estimation via quantized results}\label{ssec:vq_prior}
We extend the idea of \texttt{AllSquash} to estimate the prior probability $P(q)$ that each quantized representation occurs.
For sentence embeddings, smooth inverse frequency (SIF) \cite{arora2017simple} is often used to penalize more frequent words.
Given a word $W$ and a hyperparameter $a$, the weight $w_i = a/(a+p(W))$.
\cite{arora2017simple} points out that SIF has direct connections with TF-IDF \cite{sparck1972statistical} and the subsampling probabilities of word2vec \cite{mikolov2013linguistic}.

We extend SIF for speech representations by using VQ.
For this, we have to count the quantized indices of whole training samples and penalize the representations by their frequency:
\begin{equation}
    w^\text{SIF}_t = {a}/({a+N(q_t)}),
\end{equation}
where we denote the count of $q$ as $N(q)$ and $a$ is the smoothing hyperparameter, similar to the original SIF \cite{arora2017simple}.

However, there can be other strategies for counting, \textit{i.e.}, the criterion for the penalty $w_t$.
We developed a counting strategy that searches the indices separately in their groups, fully utilizing the group-wise similarity while avoiding sparseness \cite{baevski2020vq}:
% Function : VQ-GP/LP
\begin{equation}\label{eq:vq_prob_w}
    w^\text{GP}_t = {1}/{\textstyle\sum^{G}_{g=1}{N_g(q_t)}}
\end{equation}
where $N_g$ denotes the number of indices inside the group $g$.
We also dropped the smoothing hyperparameter $a$ for simplicity.

On the other hand, in \Cref{ssec:vq_merge}, we already suggested that similar representations may have less importance within the instance.
While we proposed the counting method that considers the global probability of VQ indices (GP), we also devised another approach that counts indices only in the instance representation.
The only difference is the scope of counting, from global to local.
We call this counting method LP.

Finally, we can use both local and global probabilities by multiplying the weights of LP and GP, denoting it BP: $w^\text{BP}_t = w^\text{LP}_t\cdot w^\text{GP}_t$, where $w^\text{LP}_t$ and $w^\text{GP}_t$ are weights of $\mathbf{C}_t$ from LP and GP, respectively.
This pooling strategy aims to reduce the weights of quantized indices that are redundant in each instance and frequently appear across the entire training set.
\section{Experimental results}\label{sec:results}
We closely follow the SUPERB benchmark that measures the efficacy of self-supervised speech models \cite{yang2021superb}, but with a key difference in the evaluation for unsupervised pooling methods (\Cref{ss:benchmark}).
For more details, consult our repository.\footnote{\url{https://github.com/IIP-Sogang/speech-pooling-benchmark}}

\subsection{Tasks} \label{ss:tasks}
We choose four speech tasks that can represent various aspects of speech processing.
% : keyword spotting, speaker identification, intent classification, and emotion recognition.
\textbf{Keyword spotting (KWS)} aims to automatically detect words within the predefined vocabulary.
We use the Google Speech Commands dataset V2 \cite{warden2018speech}, which consists of 35 keywords including numbers and words.
\textbf{Speaker identification (SID)} finds the speaker with a given utterance. 
Our benchmark focuses on text-independent closed-set SID \cite{9442674}.
We used the VoxCeleb1 dataset \cite{nagrani2017voxceleb} with the provided data splits.
\textbf{Intent classification (IC)} extracts semantic information from speech, such as the speaker's intent or emotions \cite{seo2022integration}.
We use the fluent speech command dataset\cite{lugosch2019speech} that provides labels for three different information: action, object, and location.
% Since there are 31 unique slot combinations, we consider the task as a 31-way single-label classification.
\textbf{Emotion recognition (ER)} is a task that predicts an emotion class depending on verbal channels. We adopt IEMOCAP \cite{busso2008iemocap} and select 4 emotion classes to follow \cite{yang2021superb}.
We train our models on 5-fold cross-validation using 5 sessions provided in the IEMOCAP dataset.

\subsection{Benchmark design}\label{ss:benchmark}
We design our benchmark considering the unsupervised setting: not allowing additional trainable parameters to fairly compare the effectiveness of unsupervised pooling methods. 
We evaluate the representations on various tasks by finding the nearest neighbors of each test dataset representation by measuring the distances with the utterance-level training dataset representations.
To measure the accuracy, the labels of test samples and their nearest training sample are compared, which is a commonly used setting in self-supervised representations based on contrastive losses \cite{chen2020simple,choi2022combating}.
We use the approximated nearest neighbor method \cite{bernhardsson2017annoy} to calculate distances efficiently.

\subsection{Models}
We adopted three different self-supervised speech models based on wav2vec2.0 which can generate quantized representations: \texttt{base}, \texttt{large}\cite{baevski2020wav2vec} and \texttt{XLS-R}\cite{babu2022xls-r}.
\texttt{base} and \texttt{large} differ by the model size, while \texttt{large} and \texttt{XLS-R} differ by the pretraining dataset, which is book-reading speech data and multilingual speech corpus from various sources, respectively.
We chose the widely used three models as we can assess the impact of pooling techniques under various conditions.

\subsection{Baselines}
We chose four unsupervised pooling baselines.
\textbf{Average Pooling} (AP) is the most naive and widely used baselines among various pooling methods \cite{yang2021superb}.
\textbf{Statistics Pooling} (SP) extends upon AP by utilizing not only the first-order but the second-order statistics, namely, mean and standard deviation \cite{snyder2016deep,snyder2017deep}.
\textbf{Whitening} applies the whitening transformation to the representations so that their covariance becomes an identity matrix~\cite{su2021whitening,huang2021whiteningbert}.
% We adopted it directly to the speech representations.
\textbf{SoftDecay} is the current state-of-the-art for sentence embedding, where a non-linear function is applied to make the singular value distribution of the representations less skewed~\cite{yan2022addressing}.

\begin{table}[!t]
\caption{
Unsupervised benchmark for speech representation pooling.
Test accuracy (\%) via the nearest neighbor method (\Cref{ss:benchmark})) is reported.
Best accuracies are written in bold.
}
\label{tbl:unsup}
\vspace{-0.5em}
\centering
\resizebox{0.95\columnwidth}{!}
{
\begin{tabular}{c|l|c|c|c|c|c}
%\hline
%\multicolumn{7}{c}{Testset} \\
\hline

 & Method & KWS & SID & IC & ER & AVG\\
\hline
\multirow{11}{*}{\STAB{\rotatebox[origin=c]{90}{\texttt{base}}}}
& AP & 58.44 & 13.85 & 26.71 & 53.44 & 38.11\\
& SP & 68.43 & \textbf{15.63} & 33.11 & 53.30 &42.62\\
& Whitening & 60.55 & 15.13 & 36.59 & 50.88 & 40.79\\
& SoftDecay & 58.66 & 13.66 & 26.71 & 53.44 & 38.12\\
\cline{2-7}
& VQ-Squash-AND & 58.46 & 14.82 & 26.97 & 52.76 & 38.26\\
& VQ-Squash-OR & 55.83 & 14.12 & 24.97 & 52.00 & 36.73\\
& VQ-AllSquash-AND & 65.16 & 14.80 & 32.96 & 53.50 & 41.60\\
& VQ-AllSquash-OR & 65.92 & 12.99 & 37.81 & 52.61 & 42.33\\
\cline{2-7}
& VQ-SIF & 59.62 & 13.33 & 36.54 & 54.28 & 40.94\\
& VQ-LP & \textbf{74.48} & 15.53 & 48.11 & \textbf{55.48} & \textbf{48.40}\\
& VQ-GP & 67.98 & 12.91 & 41.55 & 54.11 & 44.14\\
& VQ-BP & 73.48 & 12.94 & \textbf{50.99} & 55.13 & 48.14\\

\hline\hline
\multirow{11}{*}{\STAB{\rotatebox[origin=c]{90}{\texttt{large}}}}
& AP & 71.87 & 17.62 & 29.95 & 54.37 & 43.45\\
& SP & 77.09 & 19.15 & 36.91 & 56.77 & 47.48\\
& Whitening & 67.44 & 18.75 & 36.75 & 55.45 & 44.60\\
& SoftDecay & 72.02 & 17.37 & 30.21 & 54.37 & 43.49\\
\cline{2-7}
& VQ-Squash-AND & 70.21 & 17.05 & 31.43 & 54.12 & 43.20\\
& VQ-Squash-OR & 66.31 & 17.40 & 29.21 & 54.27 & 41.80\\
& VQ-AllSquash-AND & 76.75 & 18.77 & 39.12 & 55.34 & 47.50\\
& VQ-AllSquash-OR & 75.75 & 16.64 & 41.76 & 56.04 & 47.55\\
\cline{2-7}
& VQ-SIF & 73.09 & 17.60 & 32.64 & 54.59 & 44.48\\
& VQ-LP & \textbf{83.07} & \textbf{19.42} & 55.00 & \textbf{57.29} & \textbf{53.69}\\
& VQ-GP & 76.57 & 15.89 & 51.86 & 55.28 & 49.90\\
& VQ-BP & 81.00 & 14.71 & \textbf{58.53} & 54.28 & 52.13\\

\hline\hline
\multirow{11}{*}{\STAB{\rotatebox[origin=c]{90}{\texttt{XLS-R}}}}
& AP & 68.42 & 11.62 & 26.13 & 55.33 & 40.38\\
& SP & 77.26 & 13.42 & 36.70 & 56.54 & 45.98\\
& Whitening & 60.72 & \textbf{15.22} & 30.11 & 50.69 & 39.18\\
& SoftDecay & 68.35 & 12.02 & 26.13 & 55.33 & 40.46\\
\cline{2-7}
& VQ-Squash-AND & 70.80 & 12.06 & 34.27 & 55.35 & 43.12\\
& VQ-Squash-OR & 73.31 & 12.46 & 41.79 & 55.55 & 45.78\\
& VQ-AllSquash-AND & 75.97 & 12.82 & 44.53 & 56.72 & 47.51\\
& VQ-AllSquash-OR & 75.37 & 12.02 & 40.36 & 55.12 & 45.72\\
\cline{2-7}
& VQ-SIF & 80.45 & 11.89 & 46.08 & 57.46 & 48.97\\
& VQ-LP & 82.38 & 12.71 & 49.14 & \textbf{58.84} & 50.77\\
& VQ-GP & 82.87 & 12.00 & 52.49 & 57.74 & 51.27\\
& VQ-BP & \textbf{86.30} & 11.71 & \textbf{52.89} & 57.60 & \textbf{52.12}\\
\hline
\end{tabular}
}
\vspace{-1em}
\end{table}
\begin{figure}[t]
\centering
% \vspace{-1em}
\includegraphics[width=1.0\columnwidth]{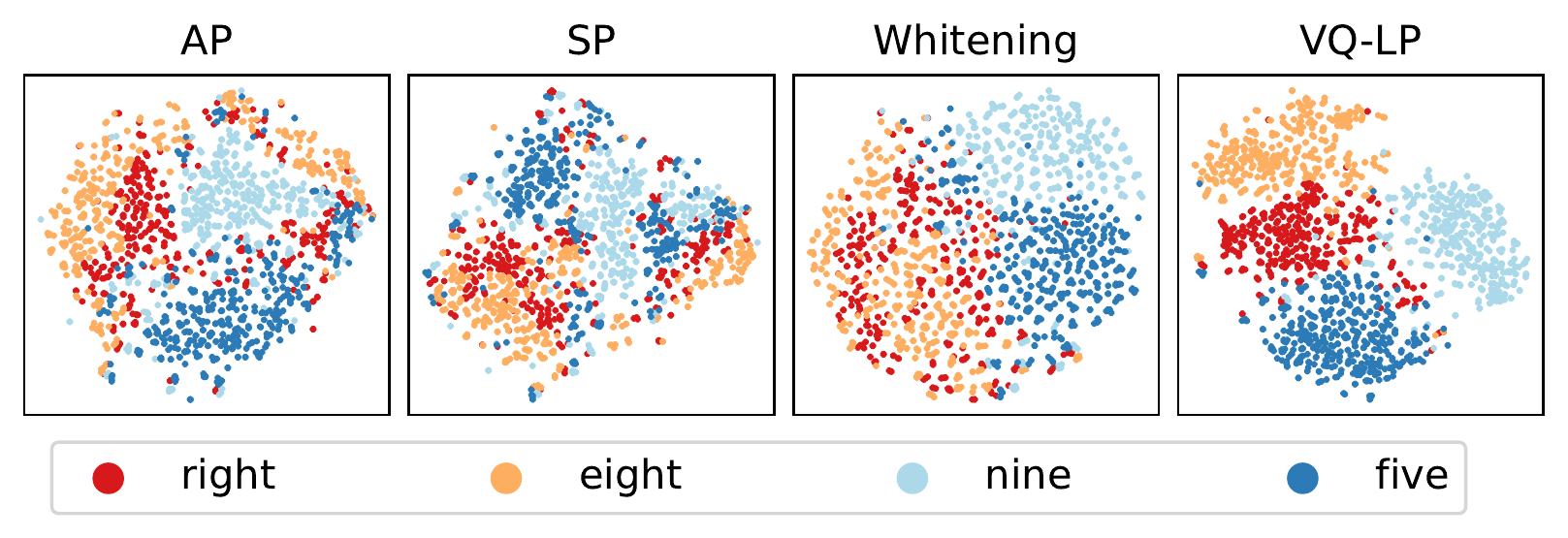}
\vspace{-2em}
\caption{
t-SNE visualization of the pooled features from Google Speech Commands V2 dataset. Each point represents a single data instance, and colors indicate four different classes.
%: ``right'', ``five'', ``nine'', and ``learn''.
}
\vspace{-2em}
\label{fig:tsne}
\end{figure}
\subsection{Results}\label{ssec:unsup_result} % Unsupervised
\Cref{tbl:unsup} shows the unsupervised benchmark results.
% \texttt{large} outperforms \texttt{base} with AP pooling, potentially due to the effectiveness of the larger model, while \texttt{XLS-R} outperforms  \texttt{large} on the ER task, leveraging the usefulness of multilingual data in capturing language-independent emotional expressions.
We found that squashing by \texttt{AND} and \texttt{OR} improved overall performance with \texttt{XLS-R} by 2.74\% and 5.4\%, respectively, compared to AP, while there was no notable improvement in other models. 
Further, the \texttt{AllSquash} strategy showed better performance than \texttt{Squash} while outperforming AP on all variants of wav2vec 2.0. 
Additionally, VQ-LP achieved the best average accuracy in \texttt{base} and \texttt{large}, while VQ-BP performed remarkably well on IC with every backbone, outscoring AP by a factor of two. 
We also employed the t-distributed Stochastic Neighbor Embedding (t-SNE) \cite{hinton2002sne} to visualize the pooled representations, as shown in \Cref{fig:tsne}.
Our observations indicate that the VQ-based approach helps differentiate between the representations of keywords with similarly pronounced vowels.
Overall, our results demonstrate the effectiveness of VQ-based pooling approaches in capturing the temporal importance of speech representations.
% We demonstrated that VQ-based pooling approaches are effective on the tested tasks, quantized indices capturing the temporal importance of speech representations to effectively pool the representations.

\begin{table}[!t]
\caption{
Performance comparison with supervised methods. Test accuracy(\%) of the best model kept by the validation set is reported. Best accuracies are written in bold.
}
\label{tbl:sup}
\centering
\resizebox{0.95\columnwidth}{!}
{
\begin{tabular}{c|l|c|c|c|c|c}
%\hline
%\multicolumn{7}{c}{Testset} \\
\hline

 & Method & KWS & SID & IC & ER & AVG\\
\hline

\multirow{6}{*}{\STAB{\rotatebox[origin=c]{90}{\texttt{base}}}}
& AP & 94.89 & 65.36 & 91.88 & 62.59 & 78.68\\
& SP & 95.15 & 64.72 & 97.39 & 55.66 & 79.18\\
& VQ-LP & 95.49 & 64.82 & 95.70 & 62.73 & 79.69\\
\cline{2-7}
& SAP & 95.15 & 40.86 & 97.39 & 55.66 & 72.27\\
& ASP & 95.70 & 36.52 & 97.47 & 59.34 & 72.26\\
& VAP & \textbf{96.74} & \textbf{68.50} & \textbf{99.50} & \textbf{65.50} & \textbf{82.56}\\
\hline\hline

\multirow{6}{*}{\STAB{\rotatebox[origin=c]{90}{\texttt{large}}}}
& AP & 96.05 & \textbf{79.23} & 95.07 & 65.47 & 83.96\\
& SP & 96.23 & 77.71 & 96.73 & 67.02 & 84.42\\
& VQ-LP & 96.32 & 78.12 & 97.76 & 66.58 & 84.70\\
\cline{2-7}
& SAP & 96.26 & 58.48 & 96.78 & 57.78 & 77.33\\
& ASP & 96.42 & 46.36 & 98.55 & 61.94 & 75.82\\
& VAP & \textbf{97.14} & 77.08 & \textbf{99.58} & \textbf{67.42} & \textbf{85.30}\\
\hline\hline

\multirow{6}{*}{\STAB{\rotatebox[origin=c]{90}{\texttt{XLS-R}}}}
& AP & 97.32 & 67.85 & 91.30 & 71.77 & 82.06 \\
& SP & 97.51 & 60.89 & 94.12 & \textbf{72.07} & 81.15\\
& VQ-LP & 97.59 & \textbf{68.13} & 94.38 & 70.67 & \textbf{82.69}\\
\cline{2-7}
& SAP & \textbf{97.63} & 61.73 & 98.86 & 64.83 & 80.76\\
& ASP & 96.69 & 13.48 & 99.02 & 61.08 & 67.57\\
& VAP & 97.49 & 61.71 & \textbf{99.63} & 71.15 & 82.49\\
\hline

\end{tabular}
}
% \vspace{-1em}
\end{table}
\begin{figure}[t]
\centering
\vspace{-1em}
\includegraphics[width=1.0\columnwidth]{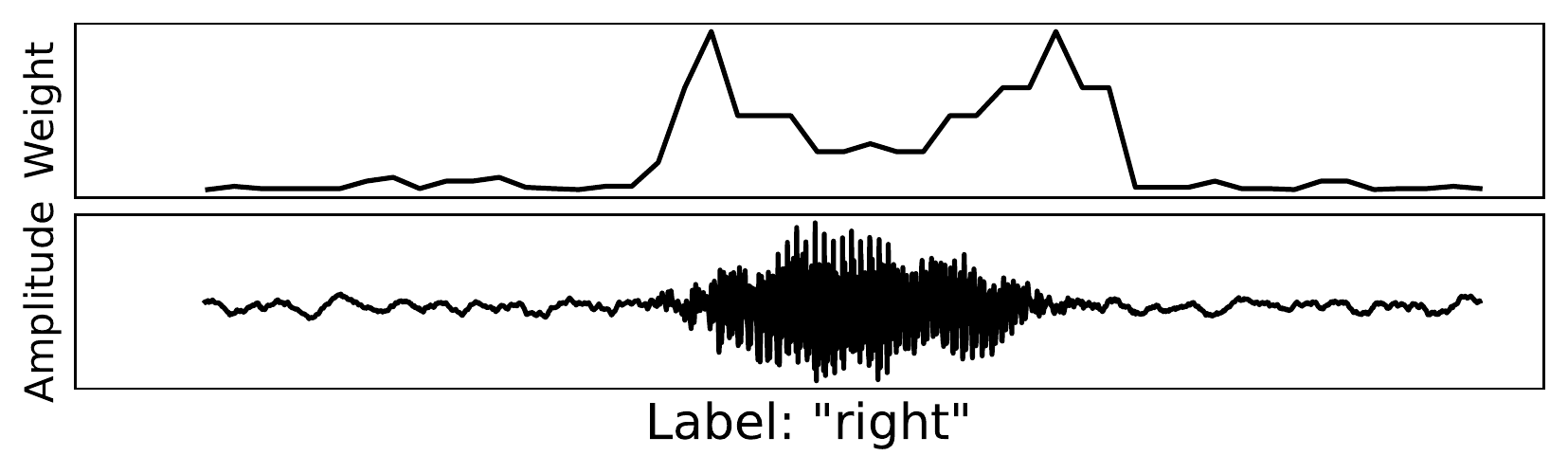}
\vspace{-2em}
\caption{
Visualized weights induced by VQ-LP. 
Two distinct peaks indicate the presence of ``r'' and ``t'' sounds, while values dropped in regions corresponding to vowel sounds. 
}
\vspace{-1em}
\label{fig:weight_wav}
\end{figure}
\section{Analysis}\label{sec:analysis}
\subsection{Comparing with parameterized methods} % supervised
In this section, we compared the effectiveness of our approach with parameterized methods.
We selected the parameterized pooling methods often used for speaker recognition: Self Attentive Pooling (SAP)\cite{Cai2018}, Attentive Statistics Pooling (ASP)\cite{okabe2018attentive}, and Vector-based Attentive Pooling (VAP)\cite{wu2020vector}.

Comparing all methods in supervised settings, we used the Adam optimizer\cite{kingma2015adam} with a default learning rate of 0.001 and 200 epochs, except for KWS using a learning rate of 0.01 and SID using 50 epochs.
Note that parameterized methods have additional trainable parameters to fit towards downstream tasks, making the comparison more favorable towards them.

While the methods using quantized representations brought a significant improvement in unsupervised experiments, the same level of enhancement was not observed in \Cref{tbl:sup}. 
However, VQ-LocalProb is comparable to the parameterized methods, even getting higher average accuracy in \texttt{XLS-R} than the best parameterized method, VAP.
The results indicate the possibility of further progress in unsupervised over supervised pooling methods, worthwhile for future investigations.

\subsection{Pooling weight analysis}
\begin{figure}[t]
\vspace{-1em}
\centering
\includegraphics[width=1.0\columnwidth]{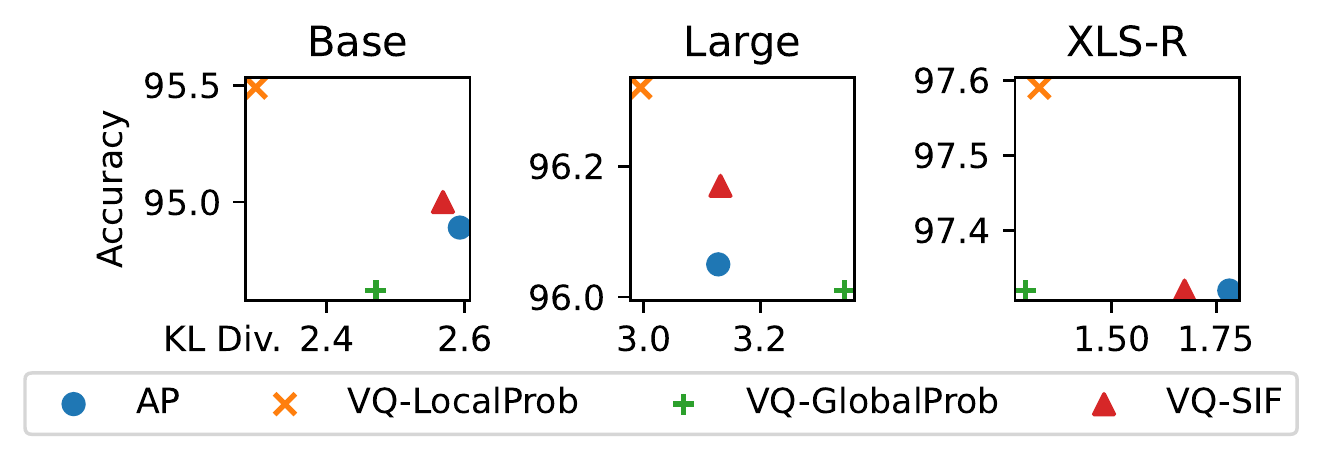}
\vspace{-2em}
\caption{
Comparing different pooling methods' weights.
KL divergence measures the similarity between the weights from SAP and others, regarding SAP as the oracle.
We observed that closer to SAP, generally better the downstream performance.
}
\vspace{-1em}
\label{fig:kl}
\end{figure}

We analyzed the behavior of our proposed pooling methods by comparing the weights of different strategies with SAP.
We use the KL divergence $D_\text{KL}(P \Vert Q)$, putting the weights induced by SAP and the others into $P$ and $Q$, respectively.
\Cref{fig:kl} show that our proposed methods generally approximate SAP better than AP.
We also found that accuracy and KL divergence negatively correlate, confirming that better similarity to SAP results in better performance. 
Additionally, we visualize a sample speech and its pooling weights induced by our VQ-LP method in \Cref{fig:weight_wav} from Speech Commands V2, where our method detects speech while suppressing areas with ambient noise.

\begin{table}[t]
\caption{
Test accuracy (\%) of average pooling on \texttt{base} with varying number of nearest neighbors (K).
}
\vspace{-1em}
\label{tbl:nnum}
\centering
\resizebox{0.95\columnwidth}{!}
{
\begin{tabular}{l|c|c|c|c|c|c|c|c}
%\hline
%\multicolumn{7}{c}{Testset} \\
\hline
\backslashbox[15mm]{Task}{K} & 1 & 3 & 4 & 5 & 6 & 7 & 8 & 9 \\
\hline
KWS & 58.44 & 60.42 & 61.88 & 62.48 & 62.58 & 62.85 & 63.22 & 63.12 \\
\hline
SID & 13.85 & 13.59 & 13.38 & 13.17 & 12.94 & 12.63 & 12.14 & 11.93 \\
\hline
IC & 26.71 & 27.58 & 27.74 & 27.42 & 28.16 & 28.16 & 28.08 & 28.05 \\
\hline
ER & 53.44 & 56.68 & 57.54 & 58.45 & 59.30 & 59.30 & 60.11 & 59.51 \\
\hline
\end{tabular}
}
\vspace{-1em}
\end{table}
\subsection{Exploring different evaluation schemes}
As the benchmark design in \Cref{ss:benchmark} utilizes only the nearest neighbor's information, we further explore the case considering a group of neighbors.
We experiment with whether increasing the number of nearest neighbors and choosing the final label via voting can increase the downstream performance. 
If a tie occurs, we follow the label of the sample that is the closest, which is the reason why we do not consider the 2-nearest case.
\Cref{tbl:nnum} shows the performance differences with respect to the number of nearest neighbors. 
Consistent improvements in performance are shown across all tasks except for SID, where the task contains relatively a large number of classes.
It indicates that the current benchmark settings can be potentially improved for more realistic use cases.
\section{Conclusion}\label{sec:conclusion}
In this paper, we design an unsupervised yet effective way of pooling variable-length speech features obtained from self-supervised speech models.
Downstream performance on various speech-processing tasks and self-supervised representations validate the effectiveness of our VQ-based approach.
Further, we compare our unsupervised pooling method with supervised pooling methods based on attention mechanisms to understand its inner workings.
Further research should be done to investigate the potential of VQ approaches on a broader array of general speech models without an embedded VQ module, such as by incorporating the idea of HuBERT \cite{hsu2021hubert}.

\bibliographystyle{IEEEtran}
\bibliography{mybib}

\end{document}